# On Inertial Navigation and Attitude Initialization in Polar Areas

Yuanxin Wu[1], Chao He[2], and Gang Liu[2]

[1]Shanghai Key Laboratory of Navigation and Location-based Services, School of Electronic Information and Electrical Engineering, Shanghai Jiao Tong University, 200240 Shanghai;

[2]Flight Automatic Control Research Institute, 710065 Xi'an.

Correspondence to: Y. X. Wu, yuanx_wu@hotmail.com

**Abstract**

Inertial navigation and attitude initialization in polar areas become a hot topic in recent years in the navigation community, as the widely-used navigation mechanization of the local level frame encounters the inherent singularity when the latitude approaches 90 degrees. Great endeavors have been devoted to devising novel navigation mechanizations such as the grid or transversal frames. This paper highlights the fact that the common Earth-frame mechanization is sufficiently good to well handle the singularity problem in polar areas. Simulation results are reported to demonstrate the singularity problem and the effectiveness of the Earth-frame mechanization.

## 1 Introduction

There are increasing demands of human activities in polar areas, such as civil aviation and underwater resource exploration. Strapdown inertial navigation systems are a kind of standard equipment for airplanes and submarines to secure an autonomous and reliable navigation means while performing those activities. Polar applications have raised a couple of challenges for inertial navigation systems.

One is the problem of attitude initialization or alignment, as the Earth gravity and the Earth rotation vector turn to be parallel near two poles, which poses a big trouble to determine the initial attitude condition for the self-contained inertial navigation systems to start with. In this regard, aiding information is necessitated to help initialize inertial navigation systems, for instance by the global navigation satellite system (GNSS). Fortunately, GNSS is able to provide reliable position and velocity information if an accurate troposphere delay model combined with dual-frequency ionosphere cancellation is used to countermeasure the problems caused by low satellite elevation (Yang and Xu 2016).

The other problem is related to the widely-used computation mechanization of the (north-pointing) local-level frame, because the north or south directions vary fast along with movement towards high-attitude regions. A partial remedy is to turn to the wander-azimuth local-level frame yet at the expense of the north direction and longitude outputs near the poles (Groves 2013). The recent years have witnessed a number of works on grid or transversal frames based mechanizations, see e.g. (Zhou, Qin et al. 2013; Li, Sun et al. 2014; Li, Ben et al. 2015; Li, Ben et al. 2016; Zhang, Yan et al. 2017; Qin, Chang et al. 2018), endeavoring to find a globally-deployable navigation mechanization to surmount the above polar singularity. However, the proposed grid or transversal navigation mechanizations have to involve to-and-fro transformations and even mechanization switches at lower-latitude areas, significantly complicating the navigation computer tasks. In fact, the current paper argues that many practitioners have neglected an obvious fact that the common Earth frame could be simply used for inertial navigation systems to achieve the global navigation ability. The work (Zhang, Yan et al. 2017) actually employs this fact yet using a concept of normal vector to encode the curvilinear position information.

**2 Attitude Alignment and Navigation Computation in Earth Frame**

In the Earth-centered Earth-fixed (ECEF) frame, the inertial navigation (attitude, velocity and position) rate equations are well-known as (Groves 2013)

$$\dot{\mathbf{C}}_b^e = \mathbf{C}_b^e \left( \boldsymbol{\omega}_{eb}^b \times \right) \tag{1}$$

$$\dot{\mathbf{v}}^e = \mathbf{C}_b^e \mathbf{f}^b - 2\boldsymbol{\omega}_{ie}^e \times \mathbf{v}^e + \mathbf{g}^e \tag{2}$$

$$\dot{\mathbf{p}}^e = \mathbf{v}^e \tag{3}$$

where $\mathbf{p}^e = \begin{bmatrix} x & y & z \end{bmatrix}^T$ denotes the ECEF coordinate of the vehicle, $\mathbf{v}^e = \begin{bmatrix} v_x & v_y & v_z \end{bmatrix}^T$ is the ground velocity expressed in the Earth frame and $\boldsymbol{\omega}_{ie}^e = \begin{bmatrix} 0 & 0 & \Omega \end{bmatrix}^T$ is the Earth rotation rate expressed in the Earth frame. $\Omega$ is the Earth rotation rate. $\mathbf{C}_e^b$ denotes the body attitude matrix from the Earth frame to the body frame, $\boldsymbol{\omega}_{eb}^b = \boldsymbol{\omega}_{ib}^b - \mathbf{C}_e^b \boldsymbol{\omega}_{ie}^e$ the body angular rate with respect to the Earth frame, $\boldsymbol{\omega}_{ib}^b$ the body angular rate measured by gyroscopes in the body frame, $\mathbf{f}^b$ the specific force measured by accelerometers in the body frame, and $\mathbf{g}^e$ the gravity vector. The $3 \times 3$ skew symmetric matrix $(\cdot \times)$ is defined so that the cross product satisfies $\mathbf{a} \times \mathbf{b} = (\mathbf{a} \times) \mathbf{b}$ for arbitrary two vectors.

## 2.1 Attitude Alignment

In general, inertial navigation systems cannot effectively accomplish attitude alignment by its own in polar areas, as the gravity and Earth rotation vectors turn to be on the same line. In order to be initialized quickly and accurately, GNSS for aviation or a doppler velocity logger for underwater vehicles is necessary (Li, Tang et al. 2013; Wu and Pan 2013). The fine alignment stage usually relies on the practical technique of extended Kalman filtering for accurate attitude as well as online calibration of inertial sensors. Again, a good initial attitude state is vital for the Kalman filtering to behave reliably. The optimization-based alignment method (Wu, Wu et al. 2011; Wu and Pan 2013; Wu, Wang et al. 2014) has been widely accepted by the community to produce a coarse but good enough initial attitude for the subsequent Kalman-filtering based fine alignment. Hereby we briefly review it in the context of Earth frame and explain how maneuvers could help speed up the GNSS-aided attitude alignment in polar applications. In reference to the work (Wu and Pan 2013), the coarse attitude alignment in the Earth frame could be formulated as

$$\mathbf{C}_b^e(0) \boldsymbol{\alpha} = \boldsymbol{\beta} \tag{4}$$

where

$$\begin{aligned}
\boldsymbol{\alpha} &\triangleq \int_0^t \mathbf{C}_{b(t)}^{b(0)} \mathbf{f}^b dt \\
\boldsymbol{\beta} &\triangleq \mathbf{C}_{e(t)}^{e(0)} \mathbf{v}^e - \mathbf{v}^e(0) + \int_0^t \mathbf{C}_{e(t)}^{e(0)} \boldsymbol{\omega}_{ie}^e \times \mathbf{v}^e dt - \int_0^t \mathbf{C}_{e(t)}^{e(0)} \mathbf{g}^e dt
\end{aligned} \tag{5}$$

The initial attitude matrix at time zero, namely $\mathbf{C}_b^e(0)$, can be determined by solving an eigenvector/eigenvalue problem, if the vector $\boldsymbol{\alpha}$ or $\boldsymbol{\beta}$ changes its direction in the time interval of interest. In the perspective of numerical computation, more significant the vector direction changes, more accurate the attitude matrix will be. Then the coarse attitude matrix at current time is to be obtained by $\mathbf{C}_b^e(t) = \mathbf{C}_{e(0)}^{e(t)} \mathbf{C}_b^e(0) \mathbf{C}_{b(t)}^{b(0)}$. A closer look tells that the vector $\boldsymbol{\alpha}$ solely depends on the outputs of inertial navigation systems and the vector $\boldsymbol{\beta}$ is determined by the GNSS position and velocity as well as the inertially apparent gravity $\mathbf{C}_{e(t)}^{e(0)} \mathbf{g}^e$. In polar areas, the apparent gravity roughly concentrates on a line and contributes little to the direction change of the vector $\boldsymbol{\beta}$. However, this shortcoming could be effectively counter-measured by the vehicle's velocity maneuvers, especially the direction-varying maneuvers.

### 2.2 Navigation Computation

The navigation computation procedure in the Earth frame roughly follows that of the local-level frame (North-Up-East), except being free of the singularity. Table I lists and compares the typical two-sample computation procedures in the Earth frame and the local-level frame (Groves 2013). Without loss of generality, the navigation computation is considered in the time interval

Table I. Two-sample Inertial Navigation Computation in Local-level Frame and Earth Frame

| | Local-level frame | Earth frame |
|---|---|---|
| Attitude Update | $\boldsymbol{\omega}_{ie}^n = [\Omega\cos L \ \Omega\sin L \ 0]^T$<br>$\boldsymbol{\omega}_{en}^n = [v_E^n/(R_E+h) \ v_E^n \tan L/(R_E+h) \ -v_N^n/(R_N+h)]^T$<br>$\boldsymbol{\sigma}_n = T(\boldsymbol{\omega}_{ie}^n + \boldsymbol{\omega}_{en}^n)$<br>$\mathbf{q}_{n_k}^{n_{k+1}} = \cos\frac{|\boldsymbol{\sigma}_n|}{2} + \frac{\boldsymbol{\sigma}_n}{|\boldsymbol{\sigma}_n|}\sin\frac{|\boldsymbol{\sigma}_n|}{2}$ | $\boldsymbol{\omega}_{ie}^e = [0 \ 0 \ \Omega]^T$<br>$\boldsymbol{\sigma}_e = T\boldsymbol{\omega}_{ie}^e$<br>$\mathbf{q}_{e_k}^{e_{k+1}} = \cos\frac{|\boldsymbol{\sigma}_e|}{2} + \frac{\boldsymbol{\sigma}_e}{|\boldsymbol{\sigma}_e|}\sin\frac{|\boldsymbol{\sigma}_e|}{2}$ |
| | $\boldsymbol{\sigma}_b = \Delta\boldsymbol{\theta}_1 + \Delta\boldsymbol{\theta}_2 + \frac{2}{3}\Delta\boldsymbol{\theta}_1 \times \Delta\boldsymbol{\theta}_2, \quad \mathbf{q}_{b_k}^{b_{k+1}} = \cos\frac{|\boldsymbol{\sigma}_b|}{2} + \frac{\boldsymbol{\sigma}_b}{|\boldsymbol{\sigma}_b|}\sin\frac{|\boldsymbol{\sigma}_b|}{2}$ | |
| | $\mathbf{C}_b^n(k+1) = \mathbf{C}_{n_k}^{n_{k+1}} \mathbf{C}_b^n(k) \mathbf{C}_{b_{k+1}}^{b_k}$ | $\mathbf{C}_b^e(k+1) = \mathbf{C}_{e_k}^{e_{k+1}} \mathbf{C}_b^e(k) \mathbf{C}_{b_{k+1}}^{b_k}$ |
| Velocity Update | $\mathbf{u}^b = \Delta\mathbf{v}_1 + \Delta\mathbf{v}_2 + \frac{1}{2}(\Delta\boldsymbol{\theta}_1 + \Delta\boldsymbol{\theta}_2) \times (\Delta\mathbf{v}_1 + \Delta\mathbf{v}_2) + \frac{2}{3}(\Delta\boldsymbol{\theta}_1 \times \Delta\mathbf{v}_2 + \Delta\mathbf{v}_1 \times \Delta\boldsymbol{\theta}_2)$ | |
| | $\mathbf{v}^n(k+1) = \mathbf{v}^n(k) + \mathbf{C}_b^n(k)\mathbf{u}^b$<br>$-T(2\boldsymbol{\omega}_{ie}^n + \boldsymbol{\omega}_{en}^n) \times \mathbf{v}^n(k) + T\mathbf{g}^n(k)$ | $\mathbf{v}^e(k+1) = \mathbf{v}^e(k) + \mathbf{C}_b^e(k)\mathbf{u}^b$<br>$-2T\boldsymbol{\omega}_{ie}^e \times \mathbf{v}^e(k) + T\mathbf{g}^e(k)$ |
| Position Update | $\mathbf{r} = T(\mathbf{v}^n(k) + \mathbf{v}^n(k+1))/2$<br>$\mathbf{p}^n(k+1) = \mathbf{p}^n(k) + \mathbf{R}_c(k)\mathbf{r}$ | $\mathbf{p}^e(k+1) = \mathbf{p}^e(k) + T(\mathbf{v}^e(k) + \mathbf{v}^e(k+1))/2$ |

$[t_k \ t_{k+1}]$, which contains two samples of gyroscopes (denoted as incremental angles $\Delta\boldsymbol{\theta}_1, \Delta\boldsymbol{\theta}_2$) and accelerometers (denoted as incremental velocities $\Delta\mathbf{v}_1, \Delta\mathbf{v}_2$) and $t_{k+1} - t_k \triangleq T$. The local curvature matrix $\mathbf{R}_c$ is a function of the current position (longitude $\lambda$, latitude $L$ and height $h$) as

$$\mathbf{R}_c = \begin{bmatrix} 0 & 0 & \dfrac{1}{(R_E + h)\cos L} \\ \dfrac{1}{R_N + h} & 0 & 0 \\ 0 & 1 & 0 \end{bmatrix} \tag{6}$$

where $R_E$ and $R_N$ are respectively the transverse radius of curvature and the meridian radius of curvature of the reference ellipsoid, depending on the current position as well. Specifically, the local curvature matrix $\mathbf{R}_c$ and the navigation frame's angular rate $\boldsymbol{\omega}_{en}^n$ will be subject to singularity problems while the latitude $L$ approaches 90 degrees. This is why the local-level mechanization cannot work properly near two poles. Note that throughout the paper the attitude matrix $\mathbf{C}_x^y$ is related to the attitude quaternion $\mathbf{q}_y^x \triangleq \begin{bmatrix} s & \boldsymbol{\eta}^T \end{bmatrix}^T$ by

$$\mathbf{C}_x^y = (s^2 - \boldsymbol{\eta}^T\boldsymbol{\eta})\mathbf{I}_3 + 2\boldsymbol{\eta}\boldsymbol{\eta}^T + 2s\boldsymbol{\eta}\times \tag{7}$$

It may be argued that the navigation information in the Earth frame is not intuitive for human operators to comprehend or act accordingly. As a matter of fact, this concern could be readily spared by simple coordinate transformation out of the navigation computation procedure, e.g., from the ECEF coordinate to the counterpart in the local-level frame. Figure 1 presents the flowchart of the navigation mechanization in the Earth frame, including the possible coordinate transformation for intuitive display.

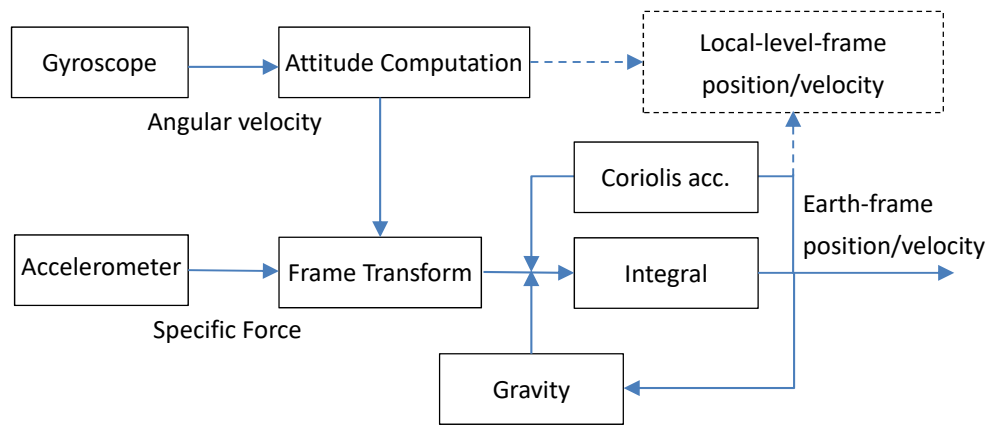

Figure 1. Earth-frame inertial navigation mechanization (with possible information display in local-level frame, as indicated by dashed lines and rectangular)

## 3 Numerical Results of Polar Navigation

This section simulates and compares the performance of the local-level and Earth-frame mechanizations in polar areas. Without loss of generality, perfect sensors and a sphere globe are considered, which does not alter the conclusions to be obtained hereafter.

Two flight scenarios have been considered, in which the flights both start from the location of longitude 120 degrees and latitude 50 degrees, a location near the Chinese city of Qiqihar. The height is kept constant at 10000 meters and the speed is also kept constant at 2000 m/s. Throughout the flight, the aircraft attitude is perfectly aligned with the Earth frame, namely $\mathbf{C}_b^e = \mathbf{I}_3$.

In the first scenario, the aircraft flies southward for an hour, arriving at latitude about minus 15 degree in the south hemisphere, while in the second scenario, it flies northward for one hour and a half, passing the north pole and finally arriving at the location of longitude 60 degrees and latitude about 33 degrees. The two flight paths are demonstrated in Fig. 2.

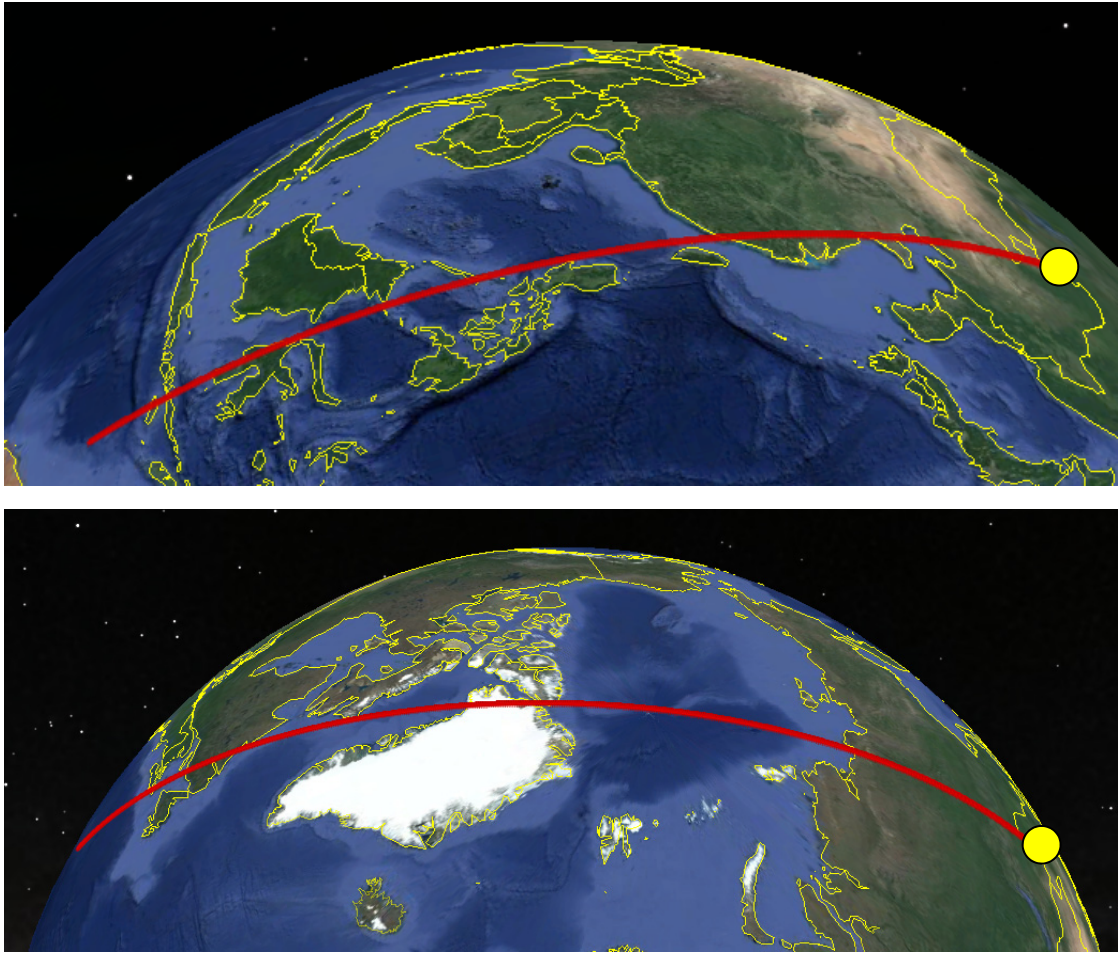

Figure 2. Two flight scenarios. (Up) southward flight; (below) northward flight passing north pole. Yellow circles indicate starting location.

The simulation data are generated analytically. Specifically, the aircraft positions in both the curvilinear and ECEF coordinates are given by

$$\mathbf{p}^n \equiv \begin{bmatrix} \lambda \\ L \\ h \end{bmatrix} = \begin{bmatrix} \lambda_0 \\ L_0 + \dfrac{v}{R+h}t \\ h_0 \end{bmatrix}, \quad \mathbf{p}^e \equiv \begin{bmatrix} x \\ y \\ z \end{bmatrix} = (R+h)\begin{bmatrix} \cos(L)\cos\lambda \\ \cos(L)\sin\lambda \\ \sin(L) \end{bmatrix} \quad (8)$$

where the initial position has $\lambda_0 = 120\deg, L_0 = 50\deg$ and $h_0 = 10000m$, and $v = 2000$ m/s. According to Eqs.(3) and (8), the velocity in the Earth frame and its derivative are obtained as

$$\mathbf{v}^e = \dot{\mathbf{p}}^e = \begin{bmatrix} -\sin(L)\cos\lambda \\ -\sin(L)\sin\lambda \\ \cos(L) \end{bmatrix} v, \quad \dot{\mathbf{v}}^e = -\begin{bmatrix} \cos(L)\cos\lambda \\ \cos(L)\sin\lambda \\ \sin(L) \end{bmatrix} \dfrac{v^2}{R+h} \quad (9)$$

Recalling $\mathbf{C}_b^e = \mathbf{I}_3$ and substituting into (2) produce the specific force measured by accelerometers $\mathbf{f}^b = \dot{\mathbf{v}}^e + 2\boldsymbol{\omega}_{ie}^e \times \mathbf{v}^e - \mathbf{g}^e$. And then according to (1), the aircraft inertial angular velocity measured by gyroscopes is $\boldsymbol{\omega}_{ib}^{b} = \boldsymbol{\omega}_{ie}^{e}$.

As the vertical channel is unstable, zero vertical velocity is assumed a priori known and used to reset the computation procedure after each update. As far as the Earth-frame mechanization is concerned, the velocity is first transformed to the local level frame and then back to the Earth frame after applying the zero vertical velocity reset.

The computation errors in the first scenario are plotted in Fig. 3. It shows that the two mechanizations perform comparably well in the non-polar regions, with a positioning error of about 20~60 meters. This error amount is negligible compared with the aviation-grade inertial navigation systems typically with a positioning error of a couple of kilometers per hour.

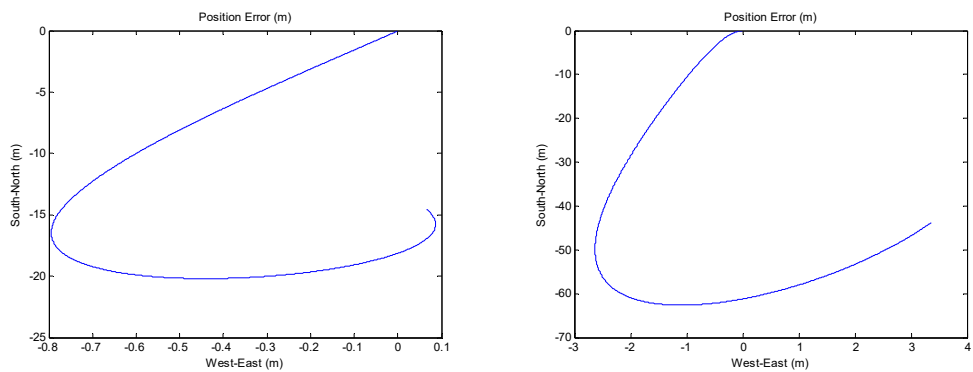

Figure 3. Non-polar position errors of local-level-frame (left) and Earth-frame mechanizations (right).

Figure 4 plots what happens in polar navigation in the second scenario. As predicated, the local-level mechanization utterly fails with its positioning error roaring to hundreds of kilometers. Meanwhile, the Earth-frame mechanization behaves in the polar region as normally as in the first scenario. The positioning error is no more than 60 meters in the whole flight.

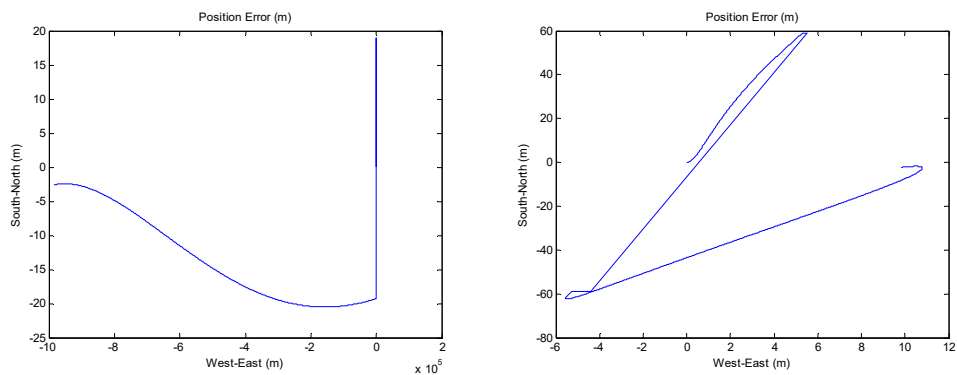

Figure 4. Polar position errors of local-level-frame (left) and Earth-frame mechanizations

(right).

Though simple, the above simulation comparison clearly is a strong support for the Earth-frame inertial navigation mechanization in various global applications.

## 4 Conclusions

Inertial navigation systems based on the widely-used local-level frame would encounter the inherent singularity problem near two poles. This paper proposes the usual but seldomly-used Earth-frame mechanization as an alternate for polar applications. Simulation results are reported in favor of this recommendation. By way of analysis, the attitude initialization with the aid of GNSS for aviation or a doppler velocity logger for underwater applications could also be safely performed in the Earth frame, even in polar regions.